\documentclass[11pt]{article} 
\usepackage
[        a4paper,
        left=2.54cm,
        right=2.54cm,
        top=2.54cm,
        bottom=2.54cm,
]
{geometry}

\usepackage{url,lineno}
\usepackage{float}
\usepackage{todonotes}
\usepackage{makecell}
\usepackage{chngpage}
\usepackage{amsmath}
\usepackage{natbib}
\usepackage{tabularx}
\usepackage{booktabs}
\usepackage{caption}
\usepackage{authblk}

\providecommand{\keywords}[1]
{
  \small	
  \textbf{\textit{Keywords }} #1
}

\begin{document}

\title{Predicting beauty, liking, and aesthetic quality: A comparative analysis of image databases for visual aesthetics research}
\author[1]{Ralf Bartho}
\author[1]{Katja Thoemmes}
\author[1]{Christoph Redies\thanks{corresponding author: christoph.redies@med.uni-jena.de}}
\affil[1]{Experimental Aesthetics Group, Institute of Anatomy I, University of Jena School of Medicine, Jena, Germany}

\maketitle
\begin{abstract}

In the fields of Experimental and Computational Aesthetics, numerous image datasets have been created over the last two decades. In the present work, we provide a comparative overview of twelve image datasets that include aesthetic ratings (beauty, liking or aesthetic quality) and investigate the reproducibility of results across different datasets. Specifically, we examine how consistently the ratings can be predicted by using either (A) a set of 20 previously studied statistical image properties, or (B) the layers of a convolutional neural network developed for object recognition. Our findings reveal substantial variation in the predictability of aesthetic ratings across the different datasets. However, consistent similarities were found for datasets containing either photographs or paintings, suggesting different relevant features in the aesthetic evaluation of these two image genres. To our surprise, statistical image properties and the convolutional neural network predict aesthetic ratings with similar accuracy, highlighting a significant overlap in the image information captured by the two methods. Nevertheless, the discrepancies between the datasets call into question the generalizability of previous research findings on single datasets. Our study underscores the importance of considering multiple datasets to improve the validity and generalizability of research results in the fields of experimental and computational aesthetics. 


\vspace{0.5cm}

\end{abstract}

\section{Introduction}

Computer vision research has benefited from the availability of large, high-quality (peer-reviewed) image datasets. Arguably, the most influential dataset in computer vision has been ImageNet \citep{ImageNet}, which was developed for visual recognition tasks. Currently, the complete ImageNet dataset contains more than 14 million images with a large variety of annotations. It is easily accessible and freely available for research. The fact that many research groups worked on the same dataset allowed for good comparability of results, a well-connected research field, high visibility, and rapid propagation of new methods, which together led to significant breakthroughs \citep{Yang2020TowardsFD}.

As obvious as the benefits of a universally accepted dataset may be, defining and creating such a dataset for aesthetic research is difficult. Compared to the classification of objects, aesthetic evaluation is a far more complex process, which is influenced not only by image features, but also by cognitive, emotional, and contextual factors (see \citealp{LederModel}, or \citealp{RediesModel}, for models of aesthetic perception). This complexity may be the reason why aesthetic research, despite many attempts, has not yet produced a universally accepted reference dataset such as ImageNet. Instead, the situation is characterized by the coexistence of many diverse datasets. 

Experimental Aesthetics and Computational Aesthetics are two major subfields of empirical aesthetics research (for a review, see \citealp{CompExpAest}). Experimental Aesthetics uses psychological research paradigms to investigate principles that underly aesthetic experience. One of its goals is to identify objective properties of visual stimuli that contribute to subjective aesthetic evaluations in the beholder. In this field, researchers often use comparatively small and specialized datasets (e.g. \citealp{ArtPics, JA, OASIS, VAPS}). Datasets often comprise hand-picked, high-resolution photographs or scans of paintings, and the aesthetic ratings are collected in laboratory experiments.
 
Computational Aesthetics uses research paradigms from computer science. Its aim is to increase the accuracy of predictions for aesthetic ratings. In this field, large and diverse datasets have been preferred (e.g. \citealp{AVA,AROD,Flickr,CB}). The images are often scraped from internet photo platforms in medium to low resolution. Most of the images are photographs and the aesthetic ratings are collected through crowd sourcing.

While the two fields of research are approaching very similar questions from different perspectives, mutually beneficial exchange between them is rather scarce. Studies in both Experimental Aesthetics and Computational Aesthetics are usually based on specifically selected (or even purpose-build) datasets adversely affecting both comparability and generalizability of results.

The datasets used in Experimental Aesthetics and Computational Aesthetics thus differ greatly in terms of number, origin and type of images. Moreover, the datasets were rated according to different aesthetic rating types, such as beauty, liking or aesthetic quality. With this in mind, the goal of this study is not to evaluate the quality of existing datasets or to recommend the use of one dataset over another. Rather, we want to gain insight into the extent to which research results can be replicated across different datasets.

The heterogeneity of the datasets used in aesthetic research also extends to the image features (or Statistical Image Properties; SIPs). Not only has a vast number of image features been studied, but the methods used to calculate the same or similar features may differ between studies. For example, methods for measuring visual complexity \citep{Geert} have been based on JPEG compression effectiveness \citep{JPG_comp}, edge density \citep{edge_complexity} or luminance and color gradients \citep{RediesPHOG}. \cite{complex_review} examined different complexity measures and found correlations between measures ranging from $\rho$=0.60 to $\rho$=0.82, raising the question of whether studies that use these different complexity measures really analyze the same visual phenomenon. Such variability is even found for very basic SIPs. For example, \cite{mallon} compute Aspect Ratio as height by width while \cite{Datta2006} and \cite{Iigaya2021} calculate it as width by height. 

In addition to classic hand-crafted image features, the use of learned features from neural networks has also increased in both Experimental Aesthetics and Computational Aesthetics. Convolutional Neural Networks (CNNs; \citealp{LeCun}) have lead to considerable advances and they now represent state-of-the-art methods in almost all classic areas of computer vision (for example, object recognition, scene segmentation, and facial recognition; \citealp{LeCunreview}). In the field of aesthetic research, they are also getting more and more attention recently. Standard CNNs trained for object recognition can be used to predict aesthetic judgments or to discriminate art from non-art, even though they have never been trained for these tasks \citep{BrachmannVar,Iigaya2021,Conwell}. All approaches have in common that they extract learned features from internal representations of neuronal networks. 

Despite the fact that neural networks are powerful and versatile tools, the complex information encoded in their layers of several thousand 'neurons' cannot be readily interpreted. By contrast, SIPs that reflect the distribution of color or luminance gradients have a relatively clear relationship to the function of human visual system. A recent study by \cite{Iigaya2021} found that low-level (objective) SIPs such as hue or saturation are more strongly represented in the lower layers of CNNs, while high-level (subjective) image features such as concreteness or dynamics tend to be found in the upper layers. 
To what extent the information captured by hand-crafted SIPs and the features learned by neural networks overlap, remains to be elucidated.

Furthermore, the methods for measuring the quality of predictions of aesthetic ratings are also not consistent across aesthetic research. Work in Computational Aesthetics often binarizes ratings and relies on classification models. These models, such as Support Vector Machines (SVM) or neural networks, are comparatively complex, have a high degree of freedom, require hyperparameter optimization and carry a high risk of overfitting. Usually, overfitting can be suppressed by additional regularization methods, which further increase the complexity of the classification models. In contrast, work in Experimental Aesthetics often models the data by using (multiple) linear regression models, which have a comparatively low degree of freedom, are easy to interpret and replicate, and do not require hyperparameter fine tuning. However, they have a much lower prediction accuracy. 

Taken together, the field of visual aesthetics is challenged by a lot of heterogeneity in (1) available image datasets with corresponding aesthetic annotations, (2) the computation of image features, and (3) modeling methods to predict aesthetic ratings by image features. This diversity leads to relatively poor comparability of research results. As a first step to find more common ground, we compare different aesthetic datasets in the present study, both in terms of aesthetic ratings, image features, and modeling. 

In the first part of the present paper, we provide an overview of the image datasets used in aesthetic research. In our study, we include twelve datasets that contain at least 500 images with corresponding aesthetic annotations for one or more aesthetic rating type (liking, beauty, aesthetic quality) and compare them based on objective image features (SIPs). To determine the reproducibility of results across the datasets, we assess to what extent the aesthetic ratings in each dataset can be predicted (A) by a set of 20 SIPs, (B) by the features in the different layers of a VGG19 convolutional neural network  \citep{VGG19} and C) by a combination of both. The 20 SIPs and the VGG19 features have been studied previously with respect to aesthetic ratings (see Section \ref{Material_Methods}). Note that we do not intend to obtain state-of-the-art prediction accuracy in this context. Consequently, to keep the measurement of the prediction accuracy as simple and reproducible as possible, we use regression models whenever possible. 

In the second part of the present paper, we shed some light onto the meaning of VGG19 layers by mapping SIPs onto them. We examine the extent to which the information captured by the 20 SIPs and the layers of the VGG19 overlap. Our results will help to understand what aspects of aesthetic perception CNNs capture internally. Last but not least, we hypothesize why CNNs can predict aesthetic ratings even though they have never been trained for this task. 

\section{Material \& Methods}\label{Material_Methods}

\subsection{Image Datasets}

In order to collect as many published image datasets with aesthetic ratings as possible, we performed an extensive literature search in the fields of Computational Aesthetics and Experimental Aesthetics and related subject areas. The literature search was supplemented by pertinent information from the homepage of the Grappa project at the University of Leuven\footnote{\url{www.grappaproject.eu/}}, and the review by \cite{ReviewData}. These two sources already listed six and seven datasets, respectively, of the twelve datasets used in the present study (Table 1). In order to be included in our study, a dataset had to meet two criteria: firstly, it had to contain a minimum of 500 images, and secondly, the rating types had to be closely associated with aesthetic preference, such as liking, aesthetic quality, or beauty. Some datasets that met these criteria were not included in the present study because they were not publicly available and their authors did not respond to our enquiries. Figure \ref{figure1} shows the five images with the highest and lowest ratings, respectively, for nine of the selected datasets. For copyright reasons, images of the AVA, EVA and Photo.net datasets cannot be shown. 

In the following sections, we will point out some of the characteristics of the twelve datasets (Table 1). Besides the respective aesthetic ratings, most datasets contain rich additional meta-information, which is not listed here because it is not directly relevant to our research question.

\begin{table}[H]

\textbf{\refstepcounter{table}\label{table1} Table \arabic{table}.}{ List of Image Datasets Analyzed in the Present Study}
\begin{adjustwidth}{-.67in}{-.67in}
\fontsize{8}{10}\selectfont
{\begin{tabular}{lclrlllr}\toprule
Dataset  	& Type$^{[1]}$  & Origin$^{[2]}$ & \# Images (x$10^{3}$)   & Method$^{[3]}$  & Rating Type$^{[4]}$ & Scale$^{[5]}$ & \# Ratings$^{[6]}$ \\\midrule
JA    		& T			& Google Art Project  & 1.6 	    & LAB     & aesthetic, beauty 	& 0 to 100*	&  $\sim$20     \\
VAPS		& T, A		& mostly Artstor	  & 1.0  	    & LAB     & liking		    	& 1 to 7   	&  20 \\
WikiEmotions& T, A		& WikiArt		  	  & 4.1  	    & CS-CF   & liking  				& -3 to 3  	&  $\geq$10	 \\
ArtPics	    & G			& AI-generated		  & 2.3 	    & LAB+OS  & liking            	& 1 to 8    &  $\sim$84	 \\
AADB        & P 		& Flickr      		  & 10.0        & CS-AMT  & aesthetic quality  			& 0 to 5    &  5		 \\
AROD 		& P			& Flickr 			  & $\sim$380.0 & UR 	    & favs per views    	& 0 to 1    &  $\sim$6868	 \\
Flickr-AES  & P 		& Flickr			  & 40.0  	    & CS-AMT  & aesthetic  			& 1 to 5   	&  5 	 \\
HiddenBeauty& P			& Flickr			  & 10.8  	    & CS-CF   & beauty            	& 1 to 5   	&  5		 \\
AVA 		& P			& DPChallenge	      & $\sim$255.0 & UR      & challenge scores     	& 1 to 10   &  $\sim$210	 \\
EVA      	& P 		& subset of AVA	      & 4.1         & OS      & AVA score, aesthetic quality  	& 0 to 10  	&  30 to 40	 \\
OASIS		& P			& Pixabay, WA, Google & 0.9 		& CS-AMT  & beauty            	& 1 to 7   	&  $\sim$189	 \\
Photo.net   & P			& PhotoNet            & 14.8 	    & UR      & aesthetic         	& 1 to 7   	&  $\sim$12	 \\\bottomrule
\end{tabular}}
\end{adjustwidth}
\captionsetup{font=scriptsize}
\caption*{P, photographs; T, traditional paintings; A, abstract paintings ; G, AI-generated images. [2] Origin or projects from which the images of the aesthetic datasets are derived. [3] UR, user ratings from the image origin; CS-CF, Crowd-Sourcing with CrowdFlower; CS-AMT, Crowd-Sourcing with Amazon Mechanical Turk; LAB, controlled laboratory experiments; OS, other forms of online surveys. [4] Note that the EVA and JA datasets are each linked to two types of aesthetic ratings, which are denoted as JA(beauty) and JA(aesth), and EVA and EVA(AVAscore), respectively, in the text. [5] Rating scales are mostly Likert scales, with the exception of continuous ratings for JA and the logarithmically calculated measure for AROD. [6] Exact or average (indicated by $\sim$) number of ratings per image. For AROD the average number of views on Flickr is listed.}

\end{table}

\subsubsection{Paintings Datasets}

The JenAesthetics dataset \citep{JA} is one of three datasets analyzed here that contain only paintings. The images originate from the Google Art Project, which provided open access to images of paintings from various public museums. When the 1629 images from the Google Art Project were selected, the main focus was on good preservation of the paintings, high resolution and good image quality. JA contains colored oil paintings from different periods of Western art. The vast majority of the paintings is representational, with few abstract artworks only. In a lab study, both beauty (JA[beauty]) and aesthetic quality (JA[aesth]) were surveyed on a continuous scale using a sliding bar.

The Vienna Art Picture System (VAPS; \citealp{VAPS}) is another dataset of paintings, which is similar to the JA dataset. It features high-resolution images of well-preserved artworks from different art periods of Western art. Furthermore, 180 of the 999 images are abstract paintings. Ratings of liking were collected for this dataset. 

The WikiArt Emotions Dataset (WikiEmotions; \citealp{Wiki}) is the third paintings dataset that we examined. As the name already indicates, it is a subset of the WikiArt dataset\footnote{\url{www.wikiart.org}} and contains images from different art periods. Liking ratings were obtained with the crowd sourcing tool CrowdFlower. Almost half of the images in the WikiEmotions dataset are abstract paintings. This is a much larger fraction than in the JA and VAPS datasets. 

\subsubsection{AI-Generated Datasets}

\begin{figure}[H]
\centering
\includegraphics[width=\textwidth]{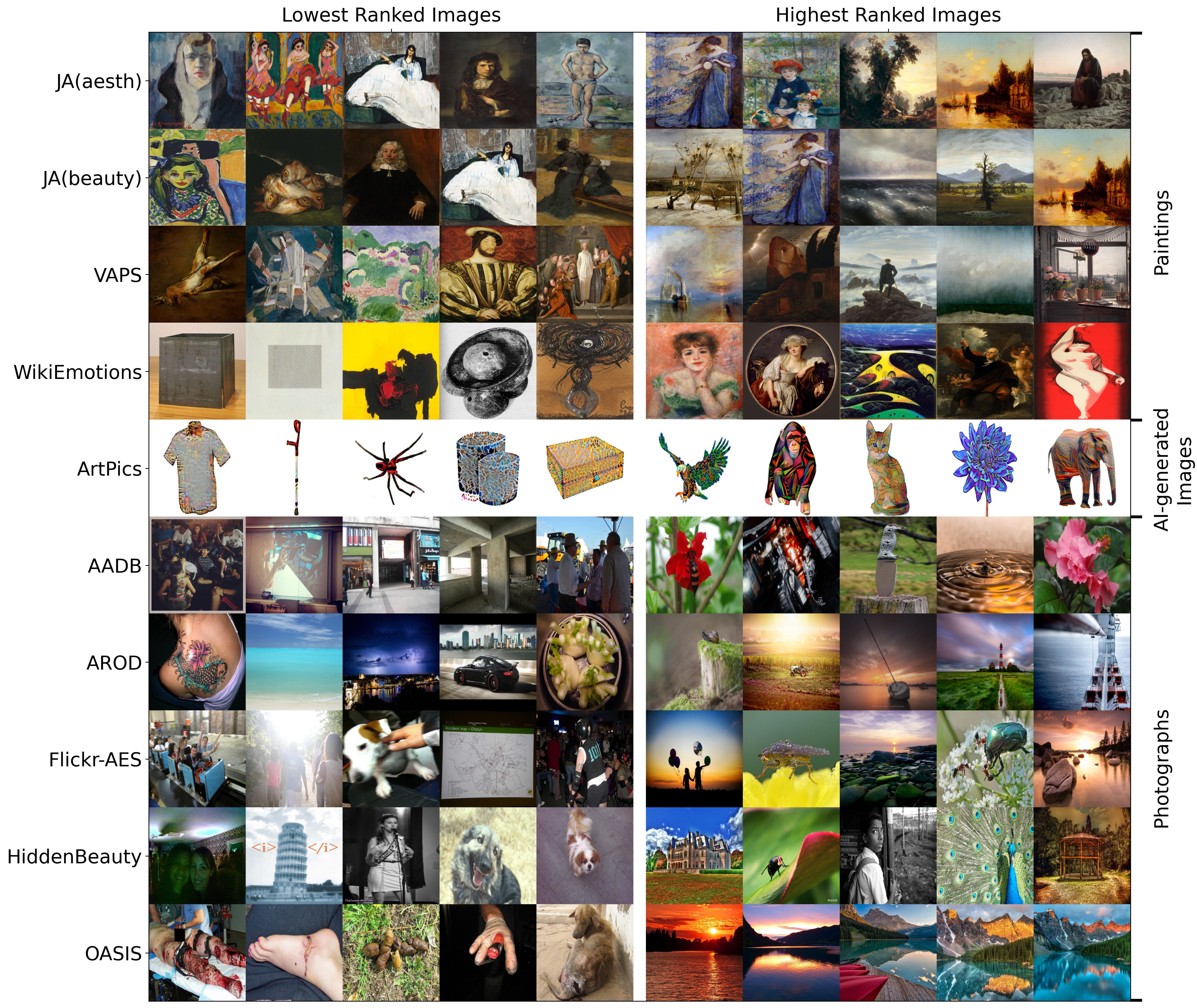}
\captionsetup{font=scriptsize}
\caption{The five images with the lowest and the highest ratings for each dataset, based on a random subset of 500 images per dataset. For copyright reasons, images of the AVA, EVA and Photo.net datasets cannot be displayed here. All images shown scaled to squares.  \label{figure1}}
\end{figure}

The ArtPics dataset \citep{ArtPics} is comparatively small and differs from the other datasets in that it is confined to AI-generated images, which show objects, plants, or animals in the center of the image on a large white background (Figure \ref{figure1}). The images were generated using Neural Style Transfer by applying the style of various paintings to the content of photographs of objects, plants, or animals. All images have the same relatively low resolution of $600 \times 450$ pixels.

\subsubsection{Photographs Datasets} 
 
The Aesthetics and Attributes Database (AADB; \citealp{AADB}) contains images from Flickr\footnote{\url{www.flickr.com}}. When selecting images for this dataset, the authors removed non-photographic images and adult content. AADB has ratings for aesthetic quality. Each image was rated by a total of 5 different participants on Amazon Mechanical Turk (AMT). The rating scale extends from 0 to 5, contrary to the information in the original paper, which mentions a scale from 1 to 5. According to one of the authors (Shu Kong, personal correspondence), this typographical mistake does not affect their results because the ratings were normalized.

The Aesthetic Ratings from Online Data dataset (AROD; \citealp{AROD}) contains images and aesthetic ratings retrieved from online information that was available on Flickr. The aesthetic measure is calculated as the ratio between the favs and the views that an image has received on Flickr. This ratio is mapped logarithmically to account for the large differences between the individual images in the number of favs and views. The rating scores of the AROD dataset thus differ from most of the other datasets analyzed, which mostly collect their ratings using explicit rating scales. In a user study, the authors showed a clear aesthetic preference for images with higher AROD scores compared to images with lower scores. Thus, their measure can be interpreted as a measure of aesthetic preference.

The Aesthetic Visual Analysis dataset (AVA; \citealp{AVA}) is one of the oldest, largest, and most widely used datasets for visual aesthetics research, especially in the field of Computational Aesthetics. The images and aesthetic ratings of the AVA dataset originate directly from the DPChallenge\footnote{\url{www.dpchallenge.com}}. At this website, an online community of photography enthusiasts rates images for photographic challenges (e.g. \textit{Animal People Interaction}, \textit{Chocolate}, \textit{Death}) on a scale from 1 to 10. Images in the AVA dataset are likely to receive high ratings when they are of particularly high aesthetic quality, meet photographic quality standards, or possibly play with the theme of the challenge in a humorous way. Thus, the reference context for the ratings is not consistent across images, which is one of the major limitations of this dataset.

The Explainable Visual Aesthetics dataset (EVA; \citealp{EVA}) is a small subset of the AVA images that contains images with an AVA rating between 4 and 9 only (original AVA scale 1-10). Furthermore, AVA images that show adult content, advertisement or a lot of text were not included. An online survey was conducted to rate the aesthetic quality of the selected images. EVA is therefore one of the few image datasets, for which two different aesthetic ratings are available (AVA score and aesthetic quality, hereafter referred to as EVA score). We analyzed the EVA images using both scores.

As the name suggests, the Flickr-AES \citep{Flickr} is another dataset that uses images from Flickr.com. In its design, Flickr-AES is almost identical to the AADB dataset. Aesthetic quality ratings were collected using AMT on a nearly identical scale of 1-5, with each image also rated by an average of 5 people. Of all the datasets examined, Flickr-AES images have the smallest median resolution (Image Size; Figure \ref{figure2}). In this respect, it differs strongly from the other datasets that also use Flickr images.

HiddenBeauty \citep{CB} is one more dataset with images from Flickr.com. It differs from the other three Flickr datasets in the rating type \textit{beauty} and the use of CrowdFlower as a survey method. In terms of size, rating scale and number of raters per image, it is very similar to AADB and Flickr-AES (Table \ref{table1}). 

With 900 images, the Open Affective Standardized Image Set (OASIS; \citealp{OASIS}) is the smallest dataset. Originally published with valence and arousal ratings, a later study by \cite{OASISratings} additionally collected ratings for \textit{beauty}, which we will use here. Nonetheless, the images were specifically selected to evoke positive and negative emotions, setting the OASIS dataset apart from others. As in the ArtPics dataset, all images in the OASIS dataset have exactly the same resolution of $500 \times 400$ pixel, which is comparatively low considering the resolution of modern desktop screens.

The online platform Photo.net was one of the first websites to provide a large number of images along with user ratings \citep{PHOTO}. The aesthetic quality of images in Photo.net was rated directly by the community of users on a scale of 1 to 7. Unlike AVA ratings, which are based on diverse photo challenges, and AROD ratings, which are calculated indirectly from favs and views, Photo.net ratings are relatively similar to surveyed ratings, even though they come from an online platform.
 
There is very little overlap between the individual datasets in the images they contain - with the exception of the EVA dataset, which is a subset of the AVA dataset. Across the four datasets derived from Flickr, we identified an exceedingly small number ($<$20) of images that are present in more than one of the datasets. A similar picture is obtained for the three datasets that contain traditional Western art (JA, VAPS, and WikiEmo). JA and VAPS share 61 images, JA and WikiEmo 77 images, and VAPS and WikiEmo 65 images. However, only a fraction of these images are exact duplicates and images of a given painting can differ substantially in saturation, contrast or resolution. 

The datasets analyzed differ widely in the number of images they contain (Table \ref{table1}). For example, the AROD dataset is 380 times larger than the VAPS dataset. These large differences could possibly introduce a strong bias in the results. To avoid effects of sample size, we carried out our analyses with a randomly selected subset of 500 images per dataset. For the large datasets in particular, a sample of 500 images might not be representative. However, it is not our intention to obtain representative results for the individual datasets, but to test the generalizability of the research results for different datasets. We thus treated each of the subsets as an independent dataset with aesthetic ratings, resulting in twelve comparable datasets. The random selection of images from each dataset ensures that the subsets include a wide range of image types and aesthetic responses. Because the JA and EVA datasets each have two different aesthetic ratings, a total of fourteen different ratings are studied. Note that the same sample of 500 images is used for each of the two JA ratings and the two EVA ratings, respectively.

The number of ratings per image also differs widely between the datasets (Table \ref{table1}). It ranges from five ratings per image for the AADB, Flickr-AES and HiddenBeauty datasets to an average of 210 ratings per image for the AVA dataset. Note that for the AROD dataset, the number 6868 denotes the number of views per image on Flickr. The low number of ratings in some databases raises questions about how representative they are, especially if rating procedures differ between datasets. \cite{AADB} report significant rater agreement (i.e. correlations) when drawing pairwise comparisons of rankings from their five raters. However, agreement among raters (i.e. shared taste) depends largely on the individual respondents as well as heterogeneity in the image material \citep{martinez2020}. For aesthetic ratings based on small sample sizes, reliability is thus very questionable. The correlation between the EVA aesthetic score and the AVA score for all images in the EVA dataset underlines this problem. The Pearson's correlation coefficient of \textit{r} = .49 (\textit{p} $<$ 0.05) is relatively low for two aesthetic ratings for the same images. This discrepancy may be due to the different survey procedures used, i.e., photographic challenges for the AVA scores \textit{versus} an online experiment with explicit rating scales for all EVA scores. Overall, when comparing different datasets, one must consider the methodological differences in data collection and the possible lack of overlap in what appear to be superficially similar aesthetic ratings.

\subsection{Statistical Image Properties - handcrafted image features}

Statistical image properties (SIPs) are objective image properties that are calculated from physical image information according to a fixed mathematical procedure and ideally capture aspects related to human vision. Examples are hue and saturation of color, or complexity and spatial distribution of luminance gradients. In Computational Aesthetics, SIPs are often referred to as handcrafted image features. Unfortunately, easy-to-use scripts or code for calculating the SIPs have not always been published alongside the scientific publications that introduce them. Therefore, it can be difficult for researchers to handle the number and variations of published SIPs. As a consequence, SIPs are not always used consistently across research groups. On this background, the 20 SIPs studied in the present paper were selected based on the following three criteria: 1) The 20 SIPs capture a wide range of visual features; 2) the SIPs have been studied in previous aesthetic research projects; and 3) the algorithms for the calculation of the SIPs have been documented in detail or we had access to the original code for the calculations. In the following paragraphs, we will briefly describe the SIPs used in the present work. For more detailed information, the reader is referred to the cited original literature (for a general review, see \cite{CompExpAest}).

\textit{Color Measures (Hue, Saturation, Luminance, Lab(a), Lab(b), Color Entropy)}. Color is a central feature of many artworks. We calculate SIPs that capture color information in HSV and CIELAB color space. Specifically, we compute the mean \textit{Hue} and the mean \textit{Saturation} in HSV space, as well as the mean values for all LAB channels (henceforth abbreviated \textit{Luminance}, \textit{Lab(a)}, \textit{Lab(b)}; note that \textit{Luminance} is also referred to as intensity in other work). As a complement to the established mean color values, we calculate the Shannon entropy of the hue channel to capture the colorfulness of an image (\textit{Color Entropy}; or \textit{HSV[H] entropy} see \cite{Geller}). This measure assumes high values if an image displays many color hues that are distributed with about equal intensity across the entire range of hues, regardless of which colors these are in detail. An image with only one color hue would have a very low Shannon entropy in the hue channel. All of these color SIPs have been previously studied in aesthetic research \citep{Datta2006,Li,mallon,ArtPics,CB,Iigaya2021,Geller}.

\textit{Aspect Ratio and Image Size}. In the present work, Image Size and Aspect Ratio are calculated as follows:

\begin{align*} 
\text{Aspect Ratio} &= \frac{\text{image width}}{\text{image height}} \\ 
\\
\text{Image Size}   &= \text{image width} + \text{image height}
\end{align*}

Although both measures are comparatively easy to calculate, some remarks are still necessary. First, the method of calculating the Aspect Ratio is not consistent in the literature. For example, \cite{mallon} calculate Aspect Ratio as height by width while \cite{Datta2006} and \cite{Iigaya2021} calculate it as width by height. In the present work, we use the latter method, because it is also the convention used to specify display formats. However, the different calculation methods have only a limited effect on the results, since both values correlate strongly with each other. Second, the image size of the images in the original dataset is not necessarily the same as the size when the ratings were collected. For some of the surveys, the images were scaled. In addition, the maximum resolution of some PC displays limit the possible image resolution. Therefore, we determined Image Size after scaling images to their actual display size, where indicated. If not specified, the original image size was used if it was displayed on a $1920 \times 1200$ pixel display without scaling. Images exceeding this resolution were scaled down (keeping the Aspect Ratio) to fit the display size. Third, we follow \cite{Datta2006} for the calculation of Image Size as the sum of image height and image width. 

\textit{Contrast and Luminance Entropy}. Contrast is a widely studied feature in aesthetic research and there are many different methods to calculate it. It is unclear to what extent these different methods capture the same visually perceivable image property \citep{CB,Tong,Li,PhotoandVideo}. In the present work, Contrast is defined as the root mean square (rms) contrast \citep{RMSContrast}, which is the standard deviation of the L channel of the CIELAB color space. We also calculate the Shannon entropy \citep{Shannon} of the L channel of the CIELAB color space. Since different entropy measures are calculated in the present work, we refer to this entropy measure as \textit{Luminance Entropy}. In other publications \citep{Sidhu,Mather,Iigaya2021}, it is often referred to simply as \textit{entropy} or \textit{Shannon entropy}.

\textit{Edge-Orientation Entropy}. Second-Order Edge-Orientation Entropy is used to measure how independently (randomly) edge orientations are distributed across an image \citep{BrachmannEdgeEntropy}. To obtain this measure, the orientation of each edge element is related to the orientation of all other edge elements in the same image by pairwise comparison. An image whose edges all have the same orientation and are distributed over the image at regular intervals would have a very low Edge-Orientation Entropy. An image with edge elements that have a random orientations and are randomly distributed over the image would have maximal Edge-Orientation Entropy. In this case, the orientations of the edge elements would be maximally independent of each other across the image.

\textit{PHOG Measures (Self-Similarity, Complexity and Anisotropy)}. Self-Similarity, Complexity and Anisotropy measures are assessed using the (Pyramid of) Histograms of Orientation Gradients ([P]HOG) method, which was originally developed for object recognition and image categorization \citep{BOSCH}. For details on the computation of Self-Similarity, Complexity, and Anisotropy, see the appendix in \cite{Braun2013}. In brief, Self-Similarity captures how similar the histograms of gradient orientations are in a pyramid of subregions of an image compared to the histogram of the entire image or other subregions. High values for \textit{Self-Similarity} indicate that the subregions are more similar to the entire image. \textit{Anisotropy} measures how different the strengths of the gradients are across orientations in an image. Lower anisotropy indicates that the strength of the oriented gradients is more uniform across orientations. Higher anisotropy means that oriented gradient strength differs more strongly. \textit{Complexity} is calculated as the mean gradient strength throughout an image. Higher complexity indicates a stronger mean gradient.

\textit{Fourier Slope and Fourier Sigma}. Fourier Slope and Fourier Sigma are based on the Fourier power spectrum of the gray-scale version of an image. Roughly speaking, the Fourier Slope indicates the relative strength of high spatial frequencies versus low spatial frequencies. The Fourier Sigma indicates how linearly the log-log plot of the Fourier spectrum decreases with increasing spatial frequency. Higher values for Fourier Sigma correspond to larger deviations from a linear course. For a more detailed description of these SIPs, see \cite{RediesFourier}. 

\textit{Symmetry-lr and Symmetry-ud}. \cite{BrachmannSymm} developed a symmetry measure that is based on the first layer of CNN filters from a pre-trained AlexNet \citep{AlexNet}. Since these filters capture both color-opponent features, luminance edges, and texture information, the symmetry measures computed from these filters more closely match the human perception of symmetry than earlier measures based on the symmetry of gray-scale pixels. For the present work, left/right symmetry (\textit{Symmetry-lr}) and up/down symmetry (\textit{Symmetry-ud}) were calculated with this method. For a broader overview of the importance and previous results on symmetry in aesthetics research, see \cite{Study_of_Symmetry}.

\textit{Sparseness and Variability of Low-Level CNN Features}. \cite{BrachmannVar} used the first convolutional layer of a pre-trained AlexNet to also measure Sparseness/Richness and Variability of the feature responses. A max-pooling operation was applied to each map of the filter responses of the 96 filters in the first CNN layer. Sparseness is calculated as the median of the variances of each max-pooling map. Variability is the variance over all entries of all max-pooling maps. Note that in the original paper by \cite{BrachmannVar}, Sparseness of SIPs was referred to as the inverse of Richness. In the present work, we decided to use the term Sparseness because the calculated variance relates directly to it (and not to its inverse value). For a visualization of Sparseness, see the boxplots in Figure \ref{figure2} for the JA dataset (traditional oil paintings; low Sparseness) compared to the ArtPics dataset (style-transferred objects on large white background; high Sparseness). 

\subsection{Image Features Learned by the VGG19 Network} 

With respect to their predictive power for aesthetic ratings, we compared the above-mentioned handcrafted features to image features learned by a CNN. In contrast to the SIPs described above, the learned features are usually high-dimensional and not easy to relate to known aspects of human vision. Because of their great predictive power, however, understanding and deciphering them promises significant advances in basic aesthetic research. 

The VGG19 is one of the most widely used CNN architectures. For example, it is the basis of many neural style transfer algorithms \citep{Gatys,STROTSS}. Based on its previous successful application in aesthetics research, we decided to use the VGG19 for the present study. VGG19 denotes the version of the VGG network with 16 convolutional and 3 fully connected layers. Like the 20 SIPs described above, we will examine the learned features of the 16 convolutional layers in terms of their predictive power with respect to the aesthetic ratings for each of the twelve datasets. We will then ask to what extent the image information captured by the SIPs and by the VGG19 overlap. Details of the procedure are described in the following section.

\subsection{Statistical Methods}\label{Stats_Meths}

Our analysis is based on subsets of the original databases with 500 randomly selected images each. An identical image number ensures good comparability of the twelve datasets when investigating the predictive power of the SIPs and the VGG19 features for the aesthetic ratings. Because the number of SIPs is 20, we decided to study the same number of the VGG19 features to ensure comparability of the results. To this aim, we use principal component analysis (PCA) to reduce the feature vectors of each VGG19 layer to the first 20 PCA components with the strongest variance. All subsequent analyses are based on 20 predictive variables, i.e. either the 20 SIPs or the 20 PCA components of the respective VGG19 layer. The only exceptions are the OASIS and ArtPics datasets, which, as mentioned earlier, contain images with fixed resolution, so the values for the SIPs Aspect Ratio and Image Size do not differ for the images. For these two datasets only, Image Size and Aspect Ratio are not included as predictor variables.

Because Experimental Aesthetics and Computational Aesthetics do not necessarily employ the same research paradigms (e.g., for significance testing, measuring predictive accuracy, or statistical tests), we use statistical methods in the present work that are standard in either field. As the regression method to model the data, we use linear regression almost exclusively. As a measure of predictive accuracy, we calculate adjusted $R^2$ values, which are closely related to the root-mean-square error that is commonly employed in Computational Aesthetics. Because both the SIPs and the ratings are not necessarily normally distributed (see Figure \ref{figure2}), we use Spearman's rank correlation coefficients to measure the relation between individual SIPs and the aesthetic ratings. To measure the overall effect size  of variables in multiple linear regression models, we calculate standardized $\beta$ values.

We also face the problem of possible multicollinearity and of other types of interferences between the 20 predictor variables. As \cite{MultiK} showed, multicollinearity between predictor variables should not necessarily be eliminated because, in some cases, it even has desirable effects. However, multicollinearity can also lead to a deterioration in predictive accuracy. In Experimental Aesthetics, the selection of the best predictor variables is usually accomplished by verifying that the variables play a consistent and interpretable role in the models and pass certain significance tests. In Computational Aesthetics, the focus is instead on generalizability of the statistical models. Here, the selection of predictor variables is often based on their predictive accuracy on a test dataset. Significance and generalizability are not necessarily interchangeable, and some statistical methods satisfy only one of the two requirements. We therefore developed a methodology for selecting predictor variables for linear regression models that is a combination of cross-validation \citep{Crossval1,Crossval2} and forward feature selection \citep{feature_sel}. 

Cross-validation is an established method in Computational Aesthetics, but it is rarely used in Experimental Aesthetics. We apply a 100-fold repeated 2-fold cross-validation in this work. The measured adjusted $R^2$ value is the mean of these 100 repetitions. This choice is motivated by the comparatively small and sometimes very heterogeneous samples of 500 images per dataset.

The forward feature selection method used in the present paper starts with an empty set of variables. First, a regression model is formed for each of the 20  variables and cross-validation is used to find the model with the highest adjusted $R^2$ value from these 20 variables. The variable corresponding to this model represents the first variable of the final set of variables. Regression models are then built again for the remaining 19 variables, with the models now consisting of two predictive variables, the best-of-20 variable selected in the previous step, and one variable from each of the remaining 19. Cross-validation is again used to determine the adjusted $R^2$ value for each of the 19 possible combinations. The variable corresponding to the best regression model is then added as a second variable to the final set of variables. Iteratively, additional variables are added to this set from the remaining variables not yet included until the adjusted $R^2$ value of the model no longer improves. This final set of variables then forms the final regression model, whose adjusted $R^2$ value is determined with cross-validation. Only for the final set of variables, the standardized $\beta$ values are determined. Note that the same combination of regression model, forward feature selection, and cross-validation is applied for the 20 SIPs, the 20 VGG19 PCA components and the combination of both.

The combination of cross-validation and forward feature selection merges paradigms from Experimental Aesthetics and Computational Aesthetics; it yields robust models with good predictive accuracy and penalizes redundant predictor variables. As a result, the remaining (reduced) subset of the original 20 predictor variables are highly significant in the regression model, which is considered essential in Experimental Aesthetics (but not so much in Computational Aesthetics). 

We also compare the 20 SIPs and the 20 VGG19 PCA components with respect to their ability to classify the content of images. For this comparison, we use a Support Vector Machine (SVM) classifier with an RBF kernel and, as in linear regression, the combination of cross-validation and forward feature selection. 

\section{Results}

This section is organized as follows: First, we provide a descriptive statistical analysis of all twelve datasets. Second, we report the Spearman coefficients of correlation between each SIP and the respective aesthetic ratings for each dataset, followed by the standardized $\beta$ values for multiple regression models based on all 20 SIPs together. We then compare the adjusted $R^2$ values of these regression models with the adjusted $R^2$ values of the VGG19 layer features and the combination of both. Finally, we give details on whether and where the SIPs are represented in the layers of the VGG19.  

\subsection{Descriptive Dataset Statistics}

\begin{figure}[H]
\centering
\includegraphics[width=\textwidth]{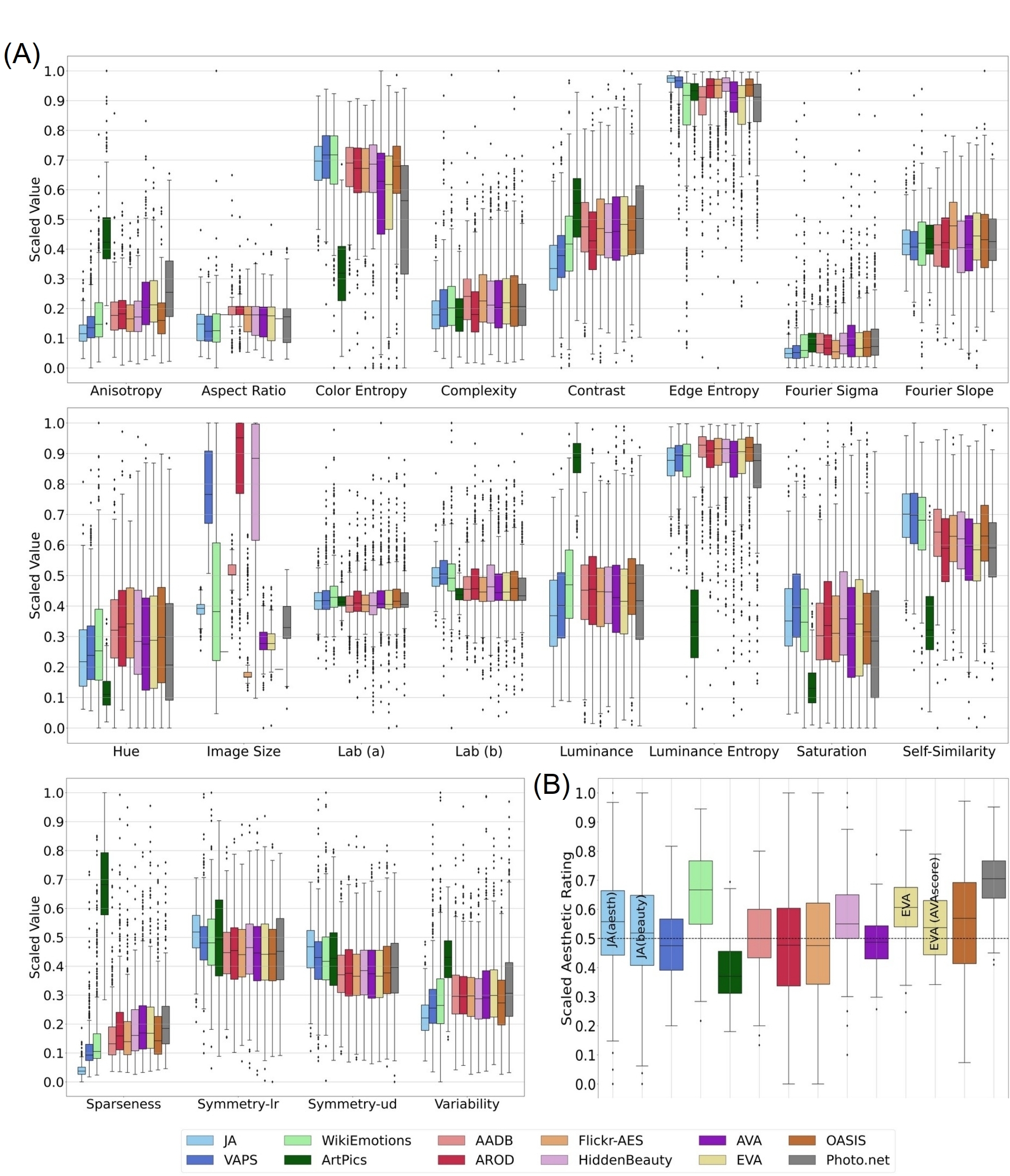}
\captionsetup{font=scriptsize}
\caption{\textbf{(A)} Boxplots of the 20 SIPs for each dataset. Values are scaled to a range of [0-1] with MinMaxScaling, i.e. for each SIP and each database, the smallest original value is scaled to 0 and the largest to 1. Note that in the ArtPics and OASIS datasets, all images have a fixed Image Size and Aspect Ratio (no boxplot). \textbf{(B)} Boxplot of the aesthetic ratings for each dataset. Values are re-scaled to a range of [0-1] using the original rating scale of the respective dataset. For example, for the OASIS dataset, which has an original rating scale of [1-7], all ratings were first reduced by -1 and then divided by 6, which moves all values into the range of [0-1] without distortion.  \label{figure2}}
\end{figure}

For each of the twelve datasets, Figure \ref{figure2}A shows boxplots of the 20 SIPs in alphabetical order. For almost all SIPs, the datasets show roughly similar patterns. An exception is Image Size, which exhibits the largest differences between the datasets. This heterogeneity reflects differences in the design and in technical limitations between the datasets. The pattern of SIPs in some of the datasets is strikingly similar. For example, the SIPs of the EVA and AVA datasets are almost identical, possibly because the EVA dataset is a subset of the AVA images. Moreover, the two datasets of traditional Western paintings, JA and VAPS, also show strong similarities (low Contrast, high Edge Entropy, low Fourier Sigma, and low Sparseness). The third dataset of paintings, WikiEmotions, stands out from the other two painting datasets by its lower Edge Entropy and significantly higher variance in Image Size.
As mentioned earlier, the AADB, Flick-AES and HiddenBeauty datasets are conceptually very similar. With the exception of Aspect Ratio, Edge Entropy and Image Size, the three datasets also show very similar distributions of their SIPs. The dataset that is most dissimilar from the others is the ArtPics dataset. It differs in almost all SIPs. This disparity is likely due to the uniformly white background in all images for the ArtPics dataset (Figure \ref{figure1}), which is not seen in the other datasets.

Figure \ref{figure2}B illustrates the distributions of the scaled aesthetic ratings for each dataset. The distributions are based on the random selection of 500 images per dataset and are thus very similar to the distributions of the respective complete datasets that were reported in the original papers, indicating that the subsets are largely representative (data not shown). A few datasets have a skewed, non-symmetric distribution of the ratings. The EVA, HiddenBeauty, JA(aesth), OASIS, Photo.net, and WikiEmotions datasets show a moderately to strongly left-skewed distribution of ratings, while the ArtPics dataset is the only one with a clear right-skewed distribution. With the exception of AROD, Flickr-AES, JA(beauty), and OASIS datasets, most datasets do not cover the full width of their rating scales and show a very peaked distribution of the ratings.

\subsection{Correlations between SIPs and ratings}

\begin{figure}[H]
\centering
\includegraphics[width=\textwidth]{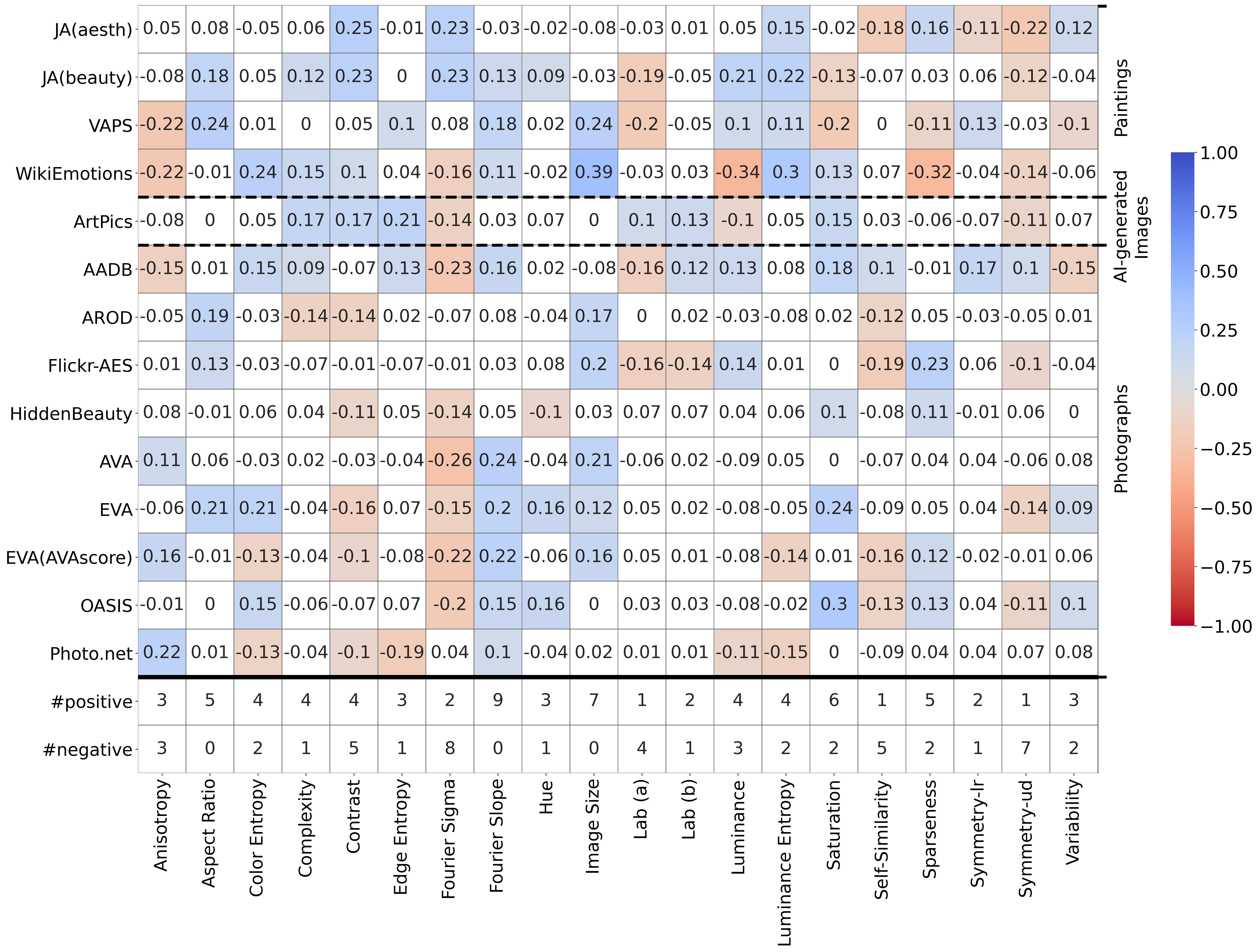}
\captionsetup{font=scriptsize}
\caption{Spearman coefficients $\rho$ for the correlation between the aesthetic ratings and the 20 SIPs for each dataset. Colored entries indicate positive (blue) and negative (red) significant correlations ($p<0.05$). Abbreviations, $\#$positive, number of positive correlations; $\#$negative, number of negative correlations. \label{figure3}}
\end{figure}

We calculated the Spearman coefficients $\rho$ for the correlation between the aesthetic ratings and each of the 20 SIPs for each dataset. Figure \ref{figure3} shows all correlations and reveals that the SIPs differ largely in the number and strength of the correlations for each dataset. All SIPs show significant correlations with a number of datasets, ranging from three datasets for Lab[b] to ten datasets for Fourier Sigma. Most SIPs show positive and negative correlations, respectively, in different datasets. There are only three SIPs (Fourier Slope, Aspect Ratio and Image Size) that are always positively correlated with the ratings. Overall, the correlations are weak to moderate with a range between $\rho = -0.34$ (Luminance in WikiEmotions) and $\rho=0.39$ (Image Size in WikiEmotions). 

Focusing on the datasets, a similarly heterogeneous picture emerges. In the AADB dataset, 14 out of 20 SIPs have a significant correlation with the human ratings, while in the AVA dataset only four SIPs correlate significantly. The pattern of correlations is rather heterogeneous even across datasets of similar image type, image origin, rating metric or survey question (Table \ref{table1}). For example, the datasets that contain images of paintings only (JA[aesth], JA[beauty], VAPS, and WikiEmotions) differ with respect to their patterns of correlations. Likewise, the datasets with Flickr images (AADB, AROD, Flickr-AES, HiddenBeauty) are very similar conceptually, but they do not show a common pattern of correlations either. Thus, the results of one dataset do not necessarily match those of another dataset, even if it is of a similar type.

\begin{figure}[H]
	\centering
	\includegraphics[width=0.8\textwidth]{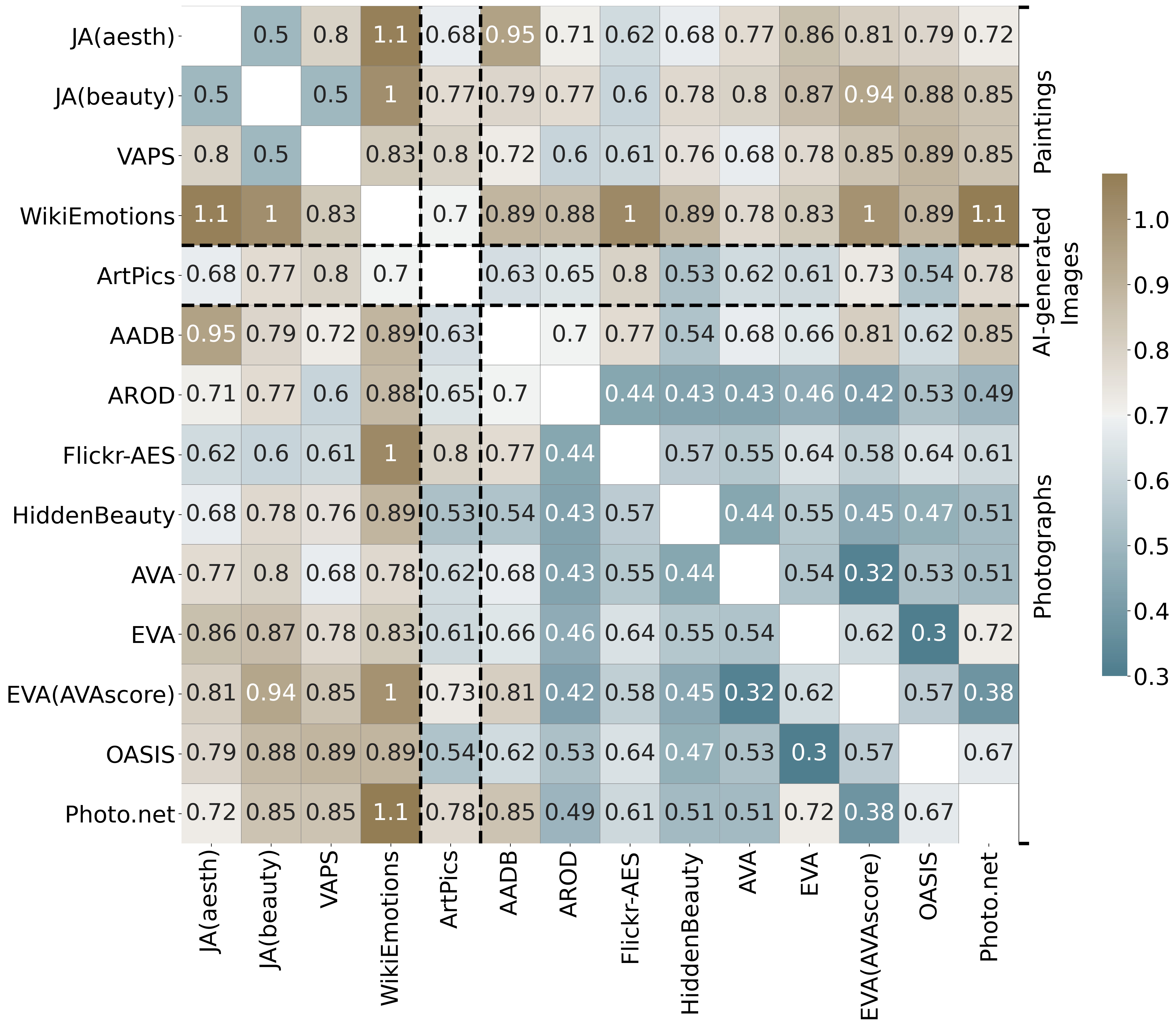}
	\captionsetup{font=scriptsize}
	\caption{ Euclidean distance of the correlation patterns between each dataset. For each row in Figure \ref{figure3} the Euclidean distance to each other row was calculated. Smaller values correspond to greater similarity between the correlation patterns of the datasets. To highlight smaller values, the color scale was centered at the median of all values. \label{figure4}}
\end{figure}

Although the datasets do not show a consistent correlation pattern overall, there are some similarities and differences between datasets containing photos vs. traditional art when doing within- and across-genre comparisons. Therefore, we analyzed the similarity of the correlation patterns in more depth. For the correlation coefficients of each row in Figure \ref{figure3}, we determined the Euclidean distance to those of each other row. This gives us a rough estimate of the similarity of the correlation patterns, with smaller distance values indicating greater similarity between the datasets. Figure \ref{figure4} shows all pairwise distances. 

Among four traditional art datasets, the patterns of correlations between the SIPs and the aesthetic ratings exhibit relatively high similarities between the two JA datasets and with the VAPS dataset (blue color in Figure \ref{figure4}). By contrast, the fourth dataset (WikiEmotions) displays low similarity in its correlation patterns with all other datasets (brown color in Figure \ref{figure4}). In contrast, within the nine photo datasets (independent of their source), there is generally high similarity, except for AADB. Comparing the art and photo datasets, it is evident that dissimilarities prevail. Nonetheless, the Flickr-AES photo dataset shows some similarity with three of the four art datasets (JA[aesth], JA[beauty], and VAPS). Interestingly, despite the ArtPics dataset (which contains AI-generated images) standing out in descriptive terms (as seen in Figure \ref{figure2}), it still displays similarities with most of the photo datasets but not so much with the art datasets.

\subsection{Multiple Linear Regression} \label{sec_betas}

\begin{figure}[H]
\centering
\includegraphics[width=\textwidth]{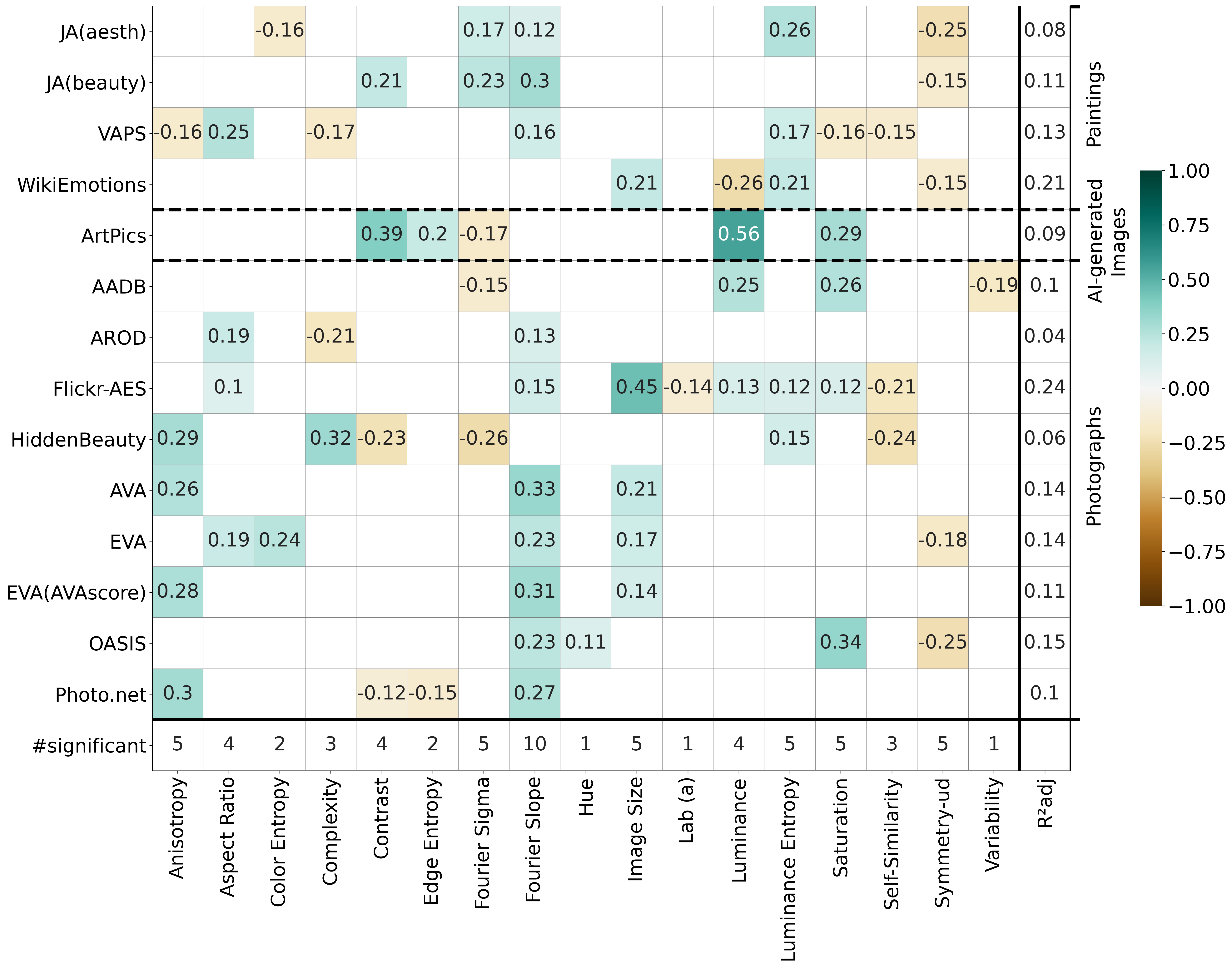}
\captionsetup{font=scriptsize}
\caption{Standardized $\beta$ values for each SIP that was selected for the multiple linear regression model of each dataset. All $\beta$ values listed are significant at $p < 0.05$. The bottom row indicates the number of times each SIP has a significant $\beta$ value ($\#$significant). The column to the right lists the adjusted $R^2$ values ($R^2 adj$) for the final models. \label{figure5}}  
\end{figure}

To eliminate possible redundancies in the predictive models due to cross-correlations between the SIPs (Suppl. Figure 1), we carried out a multiple linear regression analysis and obtained standardized $\beta$ coefficients. As described in Section \ref{Stats_Meths}, we chose a combination of cross-validation and forward feature selection to eliminate redundant and poor predictor variables that do not increase the explained variance (adjusted $R^2$) of the linear models. Figure \ref{figure5} shows the standardized $\beta$ coefficients for the SIPs selected for the final reduced model for each dataset. The colored entries indicate the SIPs that were selected as the best predictor variables during forward feature selection, all of them having a significant ($p < 0.05$) effect on the respective aesthetic ratings when the other variables were controlled for.

There are large differences between the datasets regarding the standardized $\beta$ values. The number of selected SIPs per dataset varies between three (AROD, AVA, EVA[AVAscore]) and eight (Flick-AES), suggesting that a relatively small number of the original 20 SIPs is sufficient to predict the respective aesthetic ratings in all datasets. Nevertheless, the set of selected SIPs differs between the datasets. Even when looking at conceptually similar datasets (datasets with the same image type, image origin, rating metric or survey question, Table \ref{table1}), no clear overall pattern is apparent. Interestingly, for some datasets (AROD, AVA, EVA[AVAscore], HiddenBeauty, Photo.net), the SIPs capturing color information were not selected for the best models; in these cases, the best models were built with SIPs predominantly based on grayscale versions of the images.

With regard to the frequency of the SIPs in the different models, some SIPs (Symmetry-lr , Lab[b], and Sparseness) were never selected and are therefore not shown in Figure \ref{figure5}; other SIPs were selected only once (Lab[a], Hue, and Variability). By contrast, Fourier Slope was selected for ten out of fourteen datasets. Almost half of the SIPs occur in the models with both positive and negative standardized $\beta$ coefficients. For five SIPs (Aspect Ratio, Image Size, Luminance Entropy, Fourier Slope, Symmetry-ud) $\beta$ coefficients were either all positive or all negative, respectively; these variables were also selected in at least two datasets. Notably, the $\beta$ coefficients for Fourier Slope always assume positive values. 

The last column in Figure \ref{figure5} lists the adjusted $R^2$ values for each linear model based on the forward feature selection and cross-validation. The differences between the adjusted $R^2$ values for the different datasets are large, with $R^2$ values ranging from 0.04 for the AROD dataset to 0.24 for the Flickr-AES dataset. When we clustered the datasets according to their image origin, image type, survey method, survey question, rating scale or number of Ratings per image (see Table \ref{table1}), we did not observe marked similarities within the clusters with regard to the adjusted $R^2$ values. For example, the range of the adjusted $R^2$ values is 0.08 to 0.21 for the datasets containing paintings (JA[aesth] + JA[beauty], VAPS, and WikiEmotions), and 0.04 to 0.24 for the datasets that are derived from Flickr (AADB, AROD, Flickr-AES, and HiddenBeauty).

\subsection{Predictive accuracy: Handcrafted features \textit{versus} learned features}\label{VSVSVS}

\begin{table}[H]

\textbf{\refstepcounter{table}\label{table2} Table \arabic{table}.}{ Adjusted $R^2$ values for predicting aesthetic ratings in each dataset (left-hand column) with multiple linear regression models based on a 20-dimensional PCA reduction of the VGG19 layers (L1-L16), on the 20 SIPs (SIPs), and a combination of both (SIPs \& L13).} 
\begin{adjustwidth}{-.67in}{-.67in} 
\fontsize{10}{10}\selectfont
{\begin{tabular}{l|lllllllllll|l|c}\toprule
    & \multicolumn{11}{|c|}{VGG19 layers  } & SIPs & SIPs \& L13 \\\midrule
Dataset         & L1     & L3    & L5    & L7    & L9    & L11   & L12   & L13          & L14   & L15     & L16  &  &      \\\midrule
JA(aesth)		& 0.01   & 0.03  & 0.04  & 0.04  & 0.04  & 0.06  & 0.07	& \textbf{0.11}  & 0.08  & 0.07  & 0.02  & 0.08  & 0.13  \\
JA(beauty)		& 0.03   & 0.07  & 0.08  & 0.10  & 0.08  & 0.11  & 0.09	& \textbf{0.12}  & 0.11  & 0.11  & 0.03  & 0.11  & 0.15  \\
VAPS 			& 0.06   & 0.07  & 0.05  & 0.05  & 0.06  & 0.11  & 0.13	& 0.13  & \textbf{0.15}  & 0.11  & 0.11  & 0.13  & 0.19  \\
WikiEmotions    & 0.12   & 0.09  & 0.09  & 0.13  & 0.14  & 0.18  & 0.16	& \textbf{0.21}  & 0.21  & 0.16  & 0.13  & 0.21  & 0.27  \\
ArtPics		    & 0.03   & 0.08  & 0.17  & 0.15  & 0.19  & 0.22  & 0.23	& \textbf{0.23}  & 0.18  & 0.16  & 0.14  & 0.09  & 0.24  \\
AADB        	& 0.03   & 0.05  & 0.08  & 0.09  & 0.07  & 0.09  & 0.08	& \textbf{0.10}  & 0.05  & 0.08  & 0.01  & 0.10  & 0.12  \\
AROD			& 0.02   & 0.03  & 0.03  & 0.02  & 0.03  & 0.03  & 0.03	& 0.03  & 0.03  & \textbf{0.04}  & 0.00*  & 0.04  & 0.05  \\
Flickr-AES  	& 0.04   & 0.03  & 0.03  & 0.04  & 0.08  & 0.11  & 0.11	& \textbf{0.14}  & 0.14  & 0.10  & 0.06  & 0.24  & 0.30  \\
HiddenBeauty    & 0.01   & 0.01  & 0.02  & 0.03  & 0.03  & 0.05  & 0.05	& \textbf{0.10}  & 0.07  & 0.05  & 0.04  & 0.06  & 0.13  \\
AVA 			& 0.00*  & 0.00* & 0.00* & 0.01  & 0.01  & 0.02  & 0.03	& 0.03   & 0.03  & \textbf{0.04}  & 0.03  & 0.14  & 0.16  \\
EVA      		& 0.01   & 0.03  & 0.05  & 0.06  & 0.04  & 0.09  & 0.12	& \textbf{0.12}  & 0.11  & 0.09  & 0.02  & 0.14  & 0.23  \\
EVA(AVAscores)  & 0.01   & 0.01  & 0.01  & 0.01  & 0.01  & 0.01  & 0.01 & 0.02           & 0.02  & 0.02  & \textbf{0.03} & 0.11  & 0.11  \\
OASIS			& 0.06   & 0.05  & 0.09  & 0.10  & 0.10  & 0.13  & 0.15	& 0.16  & \textbf{0.17}  & 0.10  & 0.01  & 0.15  & 0.20  \\
Photo.net     	& 0.01*  & 0.01  & 0.03  & 0.05  & 0.04  &\textbf{0.08} & 0.05	& 0.06  & 0.04  & 0.04  & 0.03   & 0.10  & 0.11  \\\midrule
mean			& 0.03	 & 0.04  & 0.06  & 0.06  &0.07   & 0.09  & 0.09 & \textbf{0.11}  & 0.10  & 0.08   & 0.05 & 0.12  & 0.17\\\bottomrule
\end{tabular}}
\end{adjustwidth}
\captionsetup{font=scriptsize}
\caption*{Bold values indicate the layer of VGG19 that has the largest adjusted $R^2$ value among all layers for the respective dataset. Asterisks indicate nonsignificant (p $>$.05) regression models. All other models were significant (p $<$.05). The last column gives the adjusted $R^2$ values for the combination of the 20 SIPs and layer 13 of the VGG19. Forward feature selection and cross-validation were performed on all regression models. Abbreviations: L1-L16, layers 1-16 of VGG19.} 

\end{table}

To directly compare the predictive power of the SIPs and the VGG19 features for aesthetic ratings, we compute adjusted $R^2$ values also for each layer of the VGG19 for each dataset, as described in Section (\ref{Stats_Meths}). Results are listed in Table \ref{table2}. Remarkably, the upper layers of VGG19 yield almost the same adjusted $R^2$ values as the 20 handcrafted SIPs. The only exceptions are the AVA, Flickr-AES and EVA(AVAscores) datasets, where SIPs predict better, and the ArtPics dataset, where layer 13 of VGG19 achieves a better prediction. In general, the adjusted $R^2$ values tend to increase throughout the layers of the VGG19, reaching a maximum (bold numbers in Table \ref{table2}) between layer 11 and layer 15 and decreasing again afterwards. This pattern is surprisingly consistent for all datasets, regardless of their image type, image origin, survey question, survey method, and rating scale.

Since the SIPs and the feature vectors of the VGG19 can predict the aesthetic ratings with similar accuracy, the question is whether they do so by using the same image information. We addressed this question by comparing the regression models using the 20 SIPs, the 20 PCA components of VGG19 layer 13, and a combination of both. Note that the same feature selection procedure (forward feature selection and cross-validation) was used in all three cases. Models based on the SIPs only had an average of 4.6 variables (Figure \ref{figure5}). The regression models that consisted of both SIPs and VGG19 PCA components had an average of 8.1 variables. Table \ref{table2} compares the adjusted $R^2$ values achieved with the values of either the SIPs or the VGG19 features alone and the combined regression models. On average, SIPs explain 12.14$\%$ of the variance (adjusted $R^2$) in aesthetic ratings.  VGG19 layer 13 explains on average 11.14$\%$. When combining the SIPs and VGG19 layer 13, the average adjusted $R^2$ increases to 17.07$\%$ explained variance. The effect is not purely additive, so there are some shared features, but the increase clearly suggests that the SIPs and the VGG19 layer 13 features also explain some unique variance. To substantiate this assumption, we next investigated whether and where SIPs are represented in the layers of VGG19.

\subsection{Mapping SIPs onto VGG19 feature layers}\label{Mapping}

\begin{figure}[H]
\centering
\includegraphics[width=\textwidth]{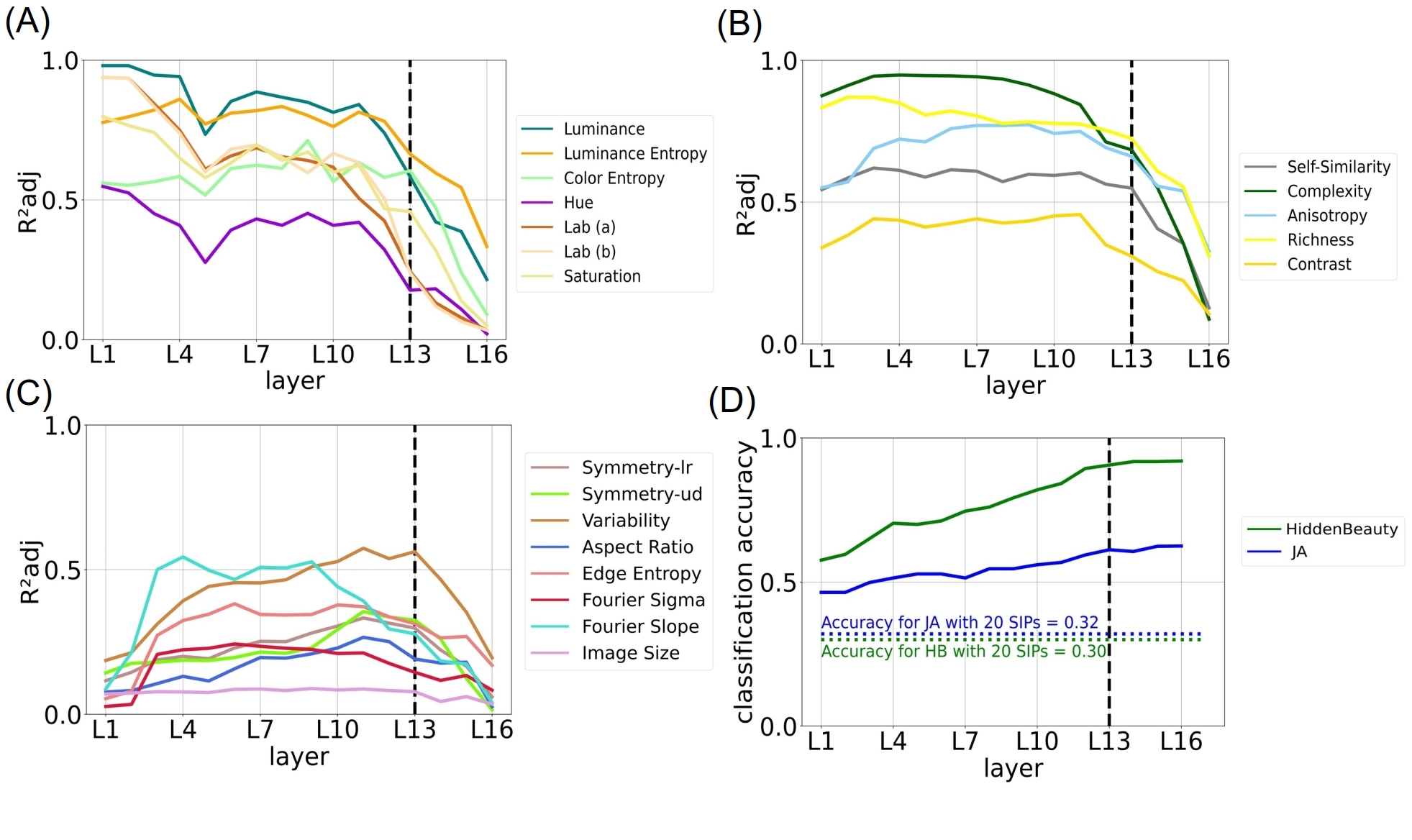}
\captionsetup{font=scriptsize}
\caption{\textbf{(A-C)} Adjusted $R^2$ values ($R^2$adj) for the prediction of the twenty SIPs by the 20 PCA components of each VGG19 layer (multiple linear regression analysis). The panels show groups of SIPs with similar patterns of predictability by the PCA components (Groups A to C). \textbf{(D)} Classification accuracy of the 20 VGG19 PCA components of each VGG19 layer, respectively, for classifying the contents of images in the JA and HiddenBeauty datasets (solid lines). The dotted lines in \textbf{(D)} show the classification accuracy for the JA and HiddenBeauty dataset when the 20 SIPs are used as predictive variables instead of the 20 PCA components. Abbreviations: L1-L16, layers 1-16 of VGG19. \label{figure6}}
\end{figure}

To study the relation between the SIPs and the VGG19 feature layers, we used the 20 PCA components of the respective convolutional layer of VGG19 to predict the individual SIPs by using multiple linear regression. Again, we applied forward feature selection and cross-validation, as described in Section \ref{Stats_Meths}, and determined the adjusted $R^2$ value of the best model.

To describe the results, we split the SIPs into three groups that each share a similar pattern of predictability by the different layers of the VGG19 (Figure \ref{figure6}A-C). Group A includes SIPs that are best predicted by the first layers of VGG19. It includes most of the SIPs that capture low-level color information, such as Hue or Saturation. In this group (Figure \ref{figure6}A), all SIPs show a marked drop of the adjusted $R^2$ values in the fifth layer. The adjusted $R^2$ values decrease steadily with increasing layer number and finally reach their minimum at the highest layer.

\begin{table}[H]
\textbf{\refstepcounter{table}\label{table3} Table \arabic{table}.}{ Categorization of SIPs according to their properties as predictive variables in the multiple linear regression models}. The relations indicate the number of SIPs for which the condition applies.
\begin{adjustwidth}{-.60in}{-.60in} 
\fontsize{10}{10}\selectfont
{\begin{tabular}{l|p{25mm}|p{25mm}|p{27mm}|p{25mm}}
&Group 1& Group 2 & Group 3 & Group 4    \\\toprule
& \makecell[l]{Edge Entropy \\ Hue \\ Lab(b) \\ Symmetry-lr}
& \makecell[l]{Complexity \\ Contrast \\ Color Entropy \\ Lab(a) \\ Luminance \\ Sparseness \\ Variability}  
& \makecell[l]{Anisotropy \\ Fourier Sigma \\ Saturation \\ Luminance Entropy \\  Symmetry-ud}
& \makecell[l]{Aspect Ratio \\ Fourier Slope \\ Image Size}    \\\midrule
\# Significant correlation with ratings  		& $\le$ 4 & $>4$     & $>4$  & $>4$   \\\midrule
\# Selected for multiple regression model  			& $\le$ 4 & $\le$ 4  & $>4$  & $>4$   \\\midrule
+/- Alternating correlation coefficients 	    & yes      & yes       & yes    & no    \\\bottomrule
\end{tabular}}{}
\end{adjustwidth}
\end{table}

Figure \ref{figure6}B shows a second group of SIPs (Group B), which consists of the three PHOG measures (Complexity, Self-Similarity and Anisotropy) as well as Sparseness and Contrast. These SIPs start with relatively high adjusted $R^2$ values in the first layer, then increase slightly to a plateau. Around layer 13, the values start to drop sharply and reach their minimum at the highest convolutional layer (L16). 

Group C consists of the remaining SIPs and follows an inverted u-shape, starting with very low adjusted $R^2$ values in the first layers, then rising steeply to reach their maximum in the middle layers (L3-L13; Figure \ref{figure6}C). Towards the last convolutional layer (L16), the adjusted $R^2$ values decrease strongly again. Nevertheless, most of these SIPs reach their maximum just before layer 13. Interestingly, Group C contains almost all of the best predictor variables such as Symmetry-ud, Aspect Ratio, Fourier Slope and Image Size (Table \ref{table3}). In contrast to the SIPs of Group A and B, the SIPs of Group C do not reach adjusted $R^2$ values above 0.60 in any layer and are therefore comparatively poorly represented in the VGG19 layers. At least for Image Size this is not surprising, since all images are scaled to a square resolution of 244$\times$244 pixel as input for the VGG19.

From these results, we conclude that the majority of the SIPs can be predicted well by the VGG19 layers, although the prediction accuracy varies strongly between the layers. Furthermore, the VGG19 layers that best predict the aesthetic ratings (around layer 13; Table \ref{table2}) differ from the layers that best represent the SIPs (Figure \ref{figure6}). Together, these results suggest that the good predictive power of the higher VGG19 layers is not based entirely on the image information reflected by the SIPs. As these higher layers were originally trained for object recognition, we additionally analyzed the ability of the VGG19 to classify content annotations for two selected datasets. In Figure \ref{figure6}D, we show the classification accuracy of each VGG19 layer for the content annotations of two datasets: The JA dataset, which is an example of a dataset of traditional art, and the HiddenBeauty dataset, which is an example of a dataset of photographs. The classification values are based on the same 20 PCA components for each layer as in the regression analysis. Again, 100-fold repeated 2-fold cross-validation was applied. Not surprisingly, the classification accuracy increases throughout the layers and reaches a maximum at the highest layer for both datasets. For comparison, the stippled lines in Figure \ref{figure6}D indicate the classification accuracy for the content annotations based on the 20 SIPs, which is much lower than that of the VGG19 layers.

\section{Discussion}

We compare twelve different image datasets of artworks or photographs. Human participants rated the images in previous studies according to a variety of aesthetic criteria, such as beauty, liking and aesthetic quality, respectively (Table \ref{table1}). In the present study, we ask how well these ratings can be predicted based on statistical image properties (SIPs) that have previously been used in visual aesthetic research (for details, see Materials $\&$ Methods section). In terms of predictive power, we compare the SIPs to the features of a convolutional neural network (CNN) that has been trained to recognize visual objects \citep{VGG19}. At first sight, our results suggest that the image datasets differ widely not only in their pictorial content, artistic style, method of production and aesthetic annotations (Table \ref{table1}), but also in how well single or multiple SIPs or CNN layers can predict the aesthetic ratings (Figures \ref{figure3}, \ref{figure5}; Table \ref{table2}). Despite this large variability, a closer look reveals some common patterns and consistencies, which will be discussed in the following sections.

\subsection{SIPS predict aesthetic ratings}

As can be expected, the median values for the different SIPs resemble each other in some datasets more than in others (Figure \ref{figure2}). Similarities seem to coincide with specific conceptual correspondences between the datasets. For example, the AVA and EVA datasets share a common origin (DPChallenge), and the JenAesthetic and VAPS datesets both represent traditional Western paintings. These similarities suggest that the SIPs capture at least some of the image characteristics of the datasets at the descriptive level. To examine these similarities in more detail, we correlated the 20 SIPs with the aesthetic ratings for each dataset. 

We started the analysis by carrying out a single regression analysis and calculated Spearman coefficients $\rho$ for the relation between all SIPs and all datasets (Figure \ref{figure3}). The number of significant correlations diverge to a large degree for the SIPs and the datasets, even for related datasets. Nevertheless, we had the impression that at least some of the datasets shared similar patterns of correlations. To quantify the (dis-)similarity of the SIP patterns in each pair of datasets, we calculated Euclidian distances between the $\rho$ values for all possible pairs of datasets (Figure \ref{figure4}). For the painting datasets, the obtained results indicate similarities in the correlation patterns between the JenAesthetics and VAPS datasets, while the third dataset (WikiEmotions) is more dissimilar. Thus, even within the painting datasets, the similarity of the correlation patterns is not entirely consistent. Whether these differences are possibly due to the higher proportion of abstract images in the WikiEmotions dataset remains to be investigated. For the nine datasets of photographs, the correlation patterns are more similar between each other in general (with the exception of the AADB dataset) but dissimilar to the painting datasets. Thus, paintings and photographs tend to represent distinct categories of images with respect to how image properties mediate aesthetic ratings by human beholders. 

In addition, some of the 20 SIPs correlated with each other to varying degrees (Suppl. Figure 1), suggesting that simple regression models with all SIPs can be partially redundant. We therefore carried out multiple linear regression with forward feature selection and cross-validation to reduce the number of predictive SIPs and their redundancies. We also studied the predictive power of the reduced models by calculating adjusted coefficients of determination ($R^2$ values). Again, results for the different datasets vary extensively with respect to the explained variance (between 4$\%$ and 24$\%$), and the number and effect size (standardized $\beta$ values) of the predictive SIPs in the reduced models (Figure \ref{figure5}). Hoewever, we noted the following consistencies.

As expected, the SIPs that are good predictors in the multiple linear regression models (Figure \ref{figure5}), also match the SIPs that correlate directly with the ratings for a given dataset in general (Figure \ref{figure3}). A striking example is Fourier Slope, which is a predictive variable in the reduced models for ten out of fourteen datasets. Moreover, the standardized $\beta$ coefficients for Fourier Slope (Figure \ref{figure5}) as well as the Spearman coefficients $\rho$ (Figure \ref{figure3}) are all positive. Fourier Slope is therefore a highly consistent predictor in a majority of datasets of both paintings and photographs. At the other extreme are SIPs that are inconsistent predictors because they contributed only once or not at all to the reduced models (Hue, Lab[a], Lab[b], Sparseness, Symmetry-lr,  and Variability). 

However, some SIPs are not included in the respective multiple linear models, although they show a relatively strong correlation with the respective ratings (for example, Fourier Sigma in the AVA dataset). This finding can be attributed to multicollinearity between individual SIPs. If two SIPs are highly correlated, they are not likely to be both included in the same model because the selection procedure penalizes models with redundant variables. Second, some SIPs are a significant component of the multiple linear models, although they do not show significant correlations with the ratings (e.g., Saturation and Luminance Entropy in Flickr-AES dataset). This is possible because correlations and standardized $\beta$ values are identical only for models that comprise one single predictor variable. Moreover, Friedman and Wall (2005) have shown that variables that do not correlate with the target variable can still explain some of the variance left by the other predictor variables in a multiple linear model.

In summary, both the correlations coefficients $\rho$ and the standardized $\beta$ values show notable differences between the individual datasets with respect to the SIPs. Thus, it is difficult to give a short answer to the question of which SIPs are the best predictors for the aesthetic ratings. With this uncertainty in mind, we classified the SIPs into the following four groups (Table \ref{table3}): SIPs in \textit{Group 1} show significant correlations for only a few datasets, and thus can be considered inconsistent predictor variables in our image datasets. These SIPs are Edge Entropy, Lab(b), and Symmetry-lr. SIPs in  \textit{Group 2} show significant correlations for single linear regression models for most datasets, but are almost never selected for the multiple linear models. This suggests that they have some predictive power, but other selected SIPs already cover the relevant information (see Suppl. Figure 1 for single linear correlations between the SIPs). These SIPs are Sparseness, Variability, Lab(a), Hue, and Color Entropy. \textit{Group 3} consists of SIPs that have significant correlations and are selected frequently for the multiple linear models, but correlation coefficients $\rho$ can be positive or negative, depending on the dataset. These SIPs are Luminance, Saturation, Contrast, Fourier Sigma, Complexity, Anisotropy, Symmetry-ud and Luminance Entropy. \textit{Group 4} comprises the strongest predictor variables. They have many significant correlations, are selected for the multiple linear models and show non-alternating (exclusively positive or negative) correlation coefficients for all relevant datasets. These SIPs are Fourier Slope, Aspect Ratio and Image Size.

While we consider it out of the scope of the present work to delve into a detailed discussion on possible reasons why each variable belongs to its respective group, we will offer possible explanations for the SIPs in Group 4. Aesthetic ratings increase as Image Size increases. Image Size, in turn, directly relates to the resolution of an image. Image resolution relates to image quality ratings in general, although image content and size can modify this relation \citep{chu2013size}. Similarly, Aspect Ratio, which represents the width-to-height ratio of an image, exhibits a positive correlation with aesthetic ratings, with the highest mean Aspect Ratio observed in the AROD dataset (1.38) and the lowest in the VAPS dataset (1.06). The Aspect Ratio of most display devices is around 1.77 (16:9), which indicates that most images fall short of this optimal ratio. Consistently, both the Spearman correlations $\rho$ and the standardized $\beta$ values show that the aesthetic ratings increase with higher aspect ratios, as the images can fill out the display more completely. Regarding Fourier Slope, previous research \citep{FSlope} has shown that slopes between -2 and -3 are preferred by human observers in random-phase images, while ratings below and above this range decrease (inverted U-shape curve). This optimal range corresponds to the range of slope values typically found in large sets of artworks \citep{RediesFourier,GrahamFourier}. In the present study, the average Fourier Slope of all datasets analyzed was -3.18, with Flickr-AES having the largest mean Fourier Slope (-3.0) and HiddenBeauty having the smallest (-3.26). However, because 93$\%$ of all images had a Fourier Slope below -2.5, our images cover mostly the ascending part of the inverted U-shaped curve (slope values below -2.5). The overall positive effect of Fourier Slope is thus compatible with these earlier results.

The SIPs that represent color information predominantly (Hue, Lab[b], Color Entropy, and Lab[a]) are relatively weak predictor variables and thus belong to Groups 1 and 2, with the exception of Saturation (Group 3). Color information is represented also by the SIPs that are based on the more global CNN features (Symmetry-ud, Symmetry-lr, Variability, and Richness) because about half of these features react in a color-specific way at the lower CNN layers \citep{BrachmannSymm,BrachmannVar}. It is out of question that color plays an important role in the aesthetic ratings, especially for paintings, but also for color photographs. Accordingly, only for five out of fourteen different aesthetic ratings, the linear models do not contain SIPs representing color. Because of the high correlations between some of the color SIPs (for example, between Color Entropy and Saturation; Suppl. Figure 1), it is difficult to pin down individual color features that mediate the aesthetic ratings in a consistent and specific way. In future studies, SIPs could be identified that more specifically cover the color information that is relevant for aesthetic ratings. We propose that SIPs that capture color information more globally, i.e. that are not based on pixel averages, may be better suited to determine the relationship between colors and aesthetic ratings.

\subsection{Partial overlap between SIPs and VGG19 features}

Previous studies demonstrated that CNNs trained for object recognition can distinguish artworks from non-art images \citep{BrachmannVar} and predict aesthetic ratings without additional fine-tuning \citep{Iigaya2021}. In the present study, we replicate these findings using a range of different image datasets. Specifically, we show that the upper layers of the VGG19 network can predict aesthetic ratings with almost the same accuracy as handcrafted SIPs. Moreover, we find that SIPs and VGG19 features only partially overlap in their ability to predict these ratings, suggesting that SIPs and VGG19 features each captures some unique information that is not captured by the other (Table \ref{table2}). Our results indicate that the higher layers of VGG19 are particularly effective in classifying image content, with the 16th layer achieving the highest classification rate (Figure \ref{figure6}D). Given this result, it is not surprising that the content information in these layers also predicts the aesthetic ratings. However, we also find that the best predictions are achieved with the 13th layer, rather than the 16th layer. We offer the following hypothesis to explain this finding: by around the 13th layer, the content classification rate is already quite high, while visual image features, which are also primarily captured by the SIPs, are still strongly represented. In contrast, these features are only weakly present in the 16th layer. As a result, the highest adjusted $R^2$ values are obtained around the 13th VGG19 layer, where both visual image features and content information are simultaneously represented. Taken together, our findings suggest that it is not just the content of an image, but also how that content is presented, that contributes to its aesthetic appeal.

\subsection{Conclusions and implications for aesthetics research}

In summary, we analyzed twelve different datasets of paintings and photographs, comprising a total of fourteen aesthetic ratings. These datasets varied in their number, origin, and type of images, the methodology used to collect aesthetic ratings, the type of aesthetic rating, the rating scale employed, and the number of ratings per image. Our results show that the datasets also differ widely in the pattern of image properties that predict their aesthetic ratings. By quantifying the similarities of the correlation patterns, we nevertheless demonstrate that the art datasets and photography datasets, respectively, are more similar to datasets of their own kind (Figure \ref{figure4}). Moreover, some SIPs (Figure \ref{figure5}; Groups 3 and 4 in Table \ref{table3}) are more consistent predictors across datasets than others (Groups 1 and 2 in Table \ref{table3}). Despite these overall consistencies, we find that even datasets that are conceptually almost identical, differ in the precise set of SIPs that best predict ratings and in the overall adjusted $R^2$ values.    

Our results give rise to the general question of how reliable and reproducible aesthetic ratings are when they were obtained on the basis of a single dataset. They highlight the possible effect of details in the collection of rating data, for example, in the selection of the stimuli, the demographics of the participants, the rating terms, the SIPs analyzed etc. Our results suggest that even small differences in the design of the experiment can have a significant impact on the results. This sensitivity places limits on the generalizability of research findings when they are based on the analysis of a single dataset only. Moreover, our selection of fourteen datasets for the present study does not represent all published image datasets with aesthetic ratings in this highly fragmented research field. We restricted our analysis to linear regression for better comparability of the predictive variables between datasets. More flexible, non-linear statistical models may possibly lead to modifications of these results.

With these caveats in mind, what implications do our results have on the study design for aesthetic research that aims to predict subjective ratings based on objective image properties? We believe that our findings should not be an incentive to restrict studies to a single datasets, but rather to include as many datasets and as many study designs as possible in the analysis. Findings that generalize well to a set of different studies can then be regarded as robust. We therefore encourage researchers to make use of the diverse existing datasets in the aesthetic community. They provide a great opportunity to put the robustness of effects to the test.

\section*{Author contributions}
RB designed the study, retrieved and analyzed the data, wrote the code and
drafted the manuscript. CR coordinated the study. CR and KT contributed to
the conception of the study and helped draft the manuscript. All
authors have approved the final article.

\section*{Disclosure/Conflict-of-Interest Statement}
The authors declare that the research was conducted in the absence of any commercial or financial relationships that could be construed as a potential conflict of interest.

\section*{Funding} This work was supported by funds from the Institute of Anatomy I, Jena University Hospital, University of Jena, Germany.

\clearpage

\bibliography{refs}

\clearpage

\setcounter{figure}{0}

\section*{Supplementary Material}

\begin{figure}[htbp]
\begin{center}
\includegraphics[width=\textwidth]{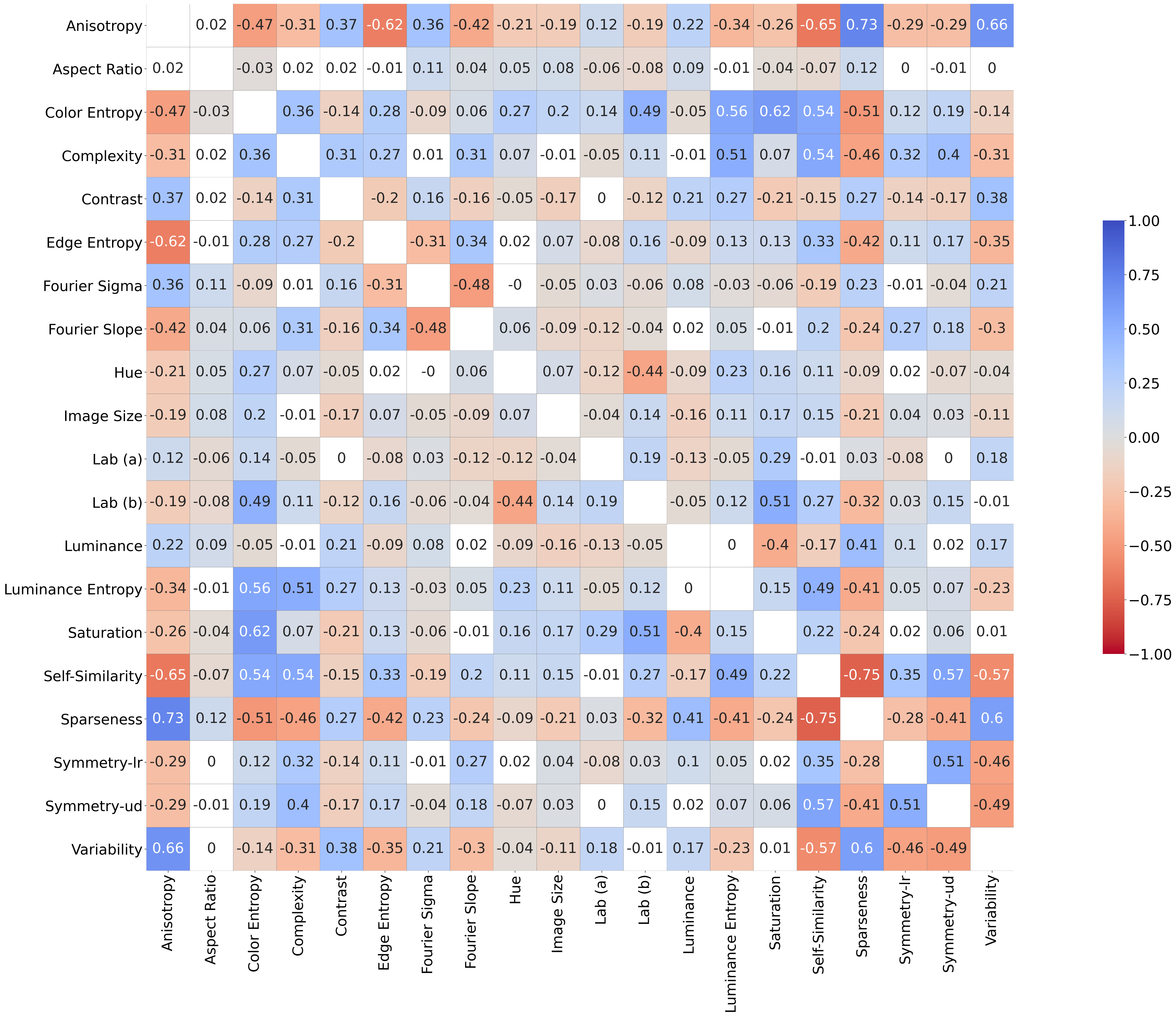}
\end{center}
\caption{Spearman coefficients $\rho$ for the correlation between the 20 SIPs for all twelve datasets combined (6000 images). Colored entries indicate positive (blue) and negative (red) significant correlations ($p<0.05$). Non-significant correlation coefficients are colored white. \label{figureS1}}
\end{figure}

\end{document}